\documentclass{article}

\usepackage[preprint]{neurips_2025}

\usepackage[utf8]{inputenc} 
\usepackage[T1]{fontenc}    
\usepackage{hyperref}       
\usepackage{url}            
\usepackage{booktabs}       
\usepackage{amsfonts}       
\usepackage{nicefrac}       
\usepackage{microtype}      
\usepackage{xcolor}         

\usepackage{graphicx}
\usepackage{wrapfig}
\usepackage{tabularx}
\usepackage{array}

\newenvironment{allintypewriter}{\ttfamily}{\par}

\title{Reasoning Large Language Model Errors Arise from Hallucinating Critical Problem Features}

\author{%
  Alex Heyman \\
  Department of Electrical Engineering \\
  and Computer Science \\
  York University \\
  Toronto, ON, Canada M3J 1P3 \\
  \texttt{aheyman@yorku.ca} \\
  \And
  Joel Zylberberg \\
  Jules Stein Eye Institute \\
  University of California \\
  Los Angeles, CA, USA 90095 \\
  \texttt{joelzy@ucla.edu} \\
}

\begin{document}

\maketitle

\begin{abstract}
Large language models have recently made great strides in reasoning task performance through chain-of-thought (CoT) strategies trained via reinforcement learning; however, these ``reasoning large language models" (RLLMs) remain imperfect reasoners, and understanding the frequencies and causes of their failure modes is important for both users and developers. We test o1-mini, o3-mini, DeepSeek-R1, Claude 3.7 Sonnet, Gemini 2.5 Pro Preview, and Grok 3 Mini Beta on graph coloring as a variable-complexity constraint-satisfaction logic problem, and find evidence from both error rate comparisons and CoT/explanation text analysis that RLLMs are prone to \textit{hallucinate graph edges} not specified in the prompt. This phenomenon persists across multiple problem complexity levels and semantic frames, and it appears to account for a significant fraction of the incorrect answers from every tested model, and the vast majority of them for some models. We also validate the generalizability of this input-conflicting hallucination phenomenon with smaller-scale experiments on a type of stable matching problem. Our results indicate that RLLMs may possess broader issues with misrepresentation of problem specifics, and we offer suggestions for design choices to mitigate this weakness.
\end{abstract}

\section{Introduction}

Large language models (LLMs) have traditionally been trained primarily through next-token prediction on large and diverse text corpuses, followed by much smaller-scale supervised fine-tuning and reinforcement learning from human feedback \citep{naveed2024comprehensive}. In the past year, however, beginning with the release of OpenAI's models o1-mini and o1-preview \citep{openai2024o1}, it has become popular for LLM developers to incorporate reinforcement learning into the training pipeline that rewards the model for producing verifiably correct answers to reasoning problems and incentivizes it to make effective use of an autoregressive natural-language ``chain of thought" (CoT). The resulting models -- sometimes called ``reasoning LLMs" (RLLMs) or variants thereof -- handily surpass traditionally trained LLMs in mathematics, coding, and other reasoning problem domains \citep{chen2025towards}, and new RLLM releases have progressively pushed the state of the art in benchmark performance.

The precise nature, capabilities, and limitations of RLLM reasoning is a subject of ongoing research, though it faces challenges both from the typical inscrutability of large neural networks and from the frequency with which developers keep model weights and/or training procedure details secret \citep{zhang2025hundred}; some model APIs even leave the chain of thought completely internal and only report the model's final answer, likely to prevent competitors from using the chains of thought as training data for their own models. Research so far has found that despite RLLMs' improvements over traditional LLMs, they are still not entirely free of flaws such as hallucinating facts in their output \citep{openai2025systemcard}, reduced performance on problems not specifically included in the training corpus \citep{mirzadeh2024gsm}, reduced performance when superficially relevant distractor information is added to the prompt \citep{mirzadeh2024gsm}, imperfect performance even on problems that are simple and algorithmically solvable without fail \citep{valmeekam2024llms, heyman2025evaluating, shojaee2025illusion}, decreasing performance with increasing problem complexity even as perfect accuracy remains algorithmically possible \citep{lin2025zebralogic, heyman2025evaluating}, and failure to correctly execute algorithms that are explicitly specified in the prompt \citep{shojaee2025illusion}. Furthermore, an RLLM's chain of thought is not guaranteed to accurately report the process by which it arrives at its final answer \citep{chen2025reasoning}. Gathering more and better evidence about the frequencies and causes of RLLM reasoning failure modes is important both for users deploying existing models and for developers seeking to build more capable and reliable models.

\begin{wrapfigure}{l}{0.5\textwidth}
\begin{center}
\includegraphics[width=0.48\textwidth]{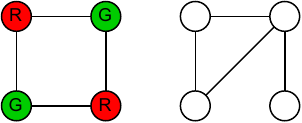}
\end{center}
\caption{\textbf{Left:} A simple undirected graph with 4 vertices and 4 edges, shown with a valid 2-coloring. \textbf{Right:} A different 4-vertex, 4-edge graph that cannot be 2-colored.}
\label{fig:graph_example}
\end{wrapfigure}

Some prior work \citep{stechly2024self, heyman2025evaluating} has evaluated LLM reasoning through \textit{graph coloring} problems, in which the model must attempt to assign one of $k$ colors to each of the $n$ vertices of a specified undirected graph such that no two vertices with an edge between them receive the same color. (See Figure \ref{fig:graph_example} for a simple example.) What colorings (if any) are valid for a given graph depends on the graph's precise arrangement of edges. A valid coloring (or proof that there is none) for a given graph and number of available colors $k$ can be found algorithmically without fail, though for $k \geq 3$ the problem is NP-complete.

In our own prior work detailed in the preprint \citet{heyman2025evaluating}, we took interest in graph coloring as a logic problem type that tests both the ability to reason step-by-step (``if I assign these colors to these vertices, which colors can I then assign to which other vertices?") and the ability to systematically explore a space of possibilities (``if assigning these colors to these vertices forces a constraint violation, what other possible color assignments can I try, or have I already tried them all?"), and for which problem instances of widely varying complexity can be easily procedurally generated. While \citet{stechly2024self} test only the traditional LLM GPT-4 on graph coloring (as well as two other problem types), we tested the RLLMs o1-mini and DeepSeek-R1 \citep{deepseek2025r1} alongside GPT-4o and three other traditional LLMs. We found that while the RLLMs perform much better than the traditional LLMs, the RLLMs remain imperfect even on simple 4-vertex 2-coloring problems, and their error rates increase dramatically as problem complexity increases to 8-vertex 4-coloring (the highest level we tested). In particular, at the 8-vertex level, both RLLMs' errors mostly consist of claiming that graphs are uncolorable under the constraints when in reality valid colorings exist (as opposed to providing colorings that are in fact invalid, either on the dataset's uncolorable problems or on its colorable ones). We also found a strong positive correlation between error rate and a measure of the difficulty of narrowing down a valid solution to a colorable problem, and we initially took this as evidence that possibility space exploration is RLLMs' primary area of weakness in the context of problems like graph coloring. However, later validation experiments that we conducted yielded evidence that this correlation is not causally significant (see Appendix \ref{app:greedy_score} for details), and thus the true primary causes of these RLLM errors likely lie elsewhere.

In this work, we investigate the failure modes of RLLMs on graph coloring problems using not only comparison of error rates and types in the models' final answers as in \citet{heyman2025evaluating}, but also analysis of the justifications they provide for those answers and of their chains of thought when available. Our experiments include o1-mini and DeepSeek-R1 as well as several newer RLLMs with superior overall benchmark performance. Remarkably, we find that all of the RLLMs we test are prone to \textit{hallucinate edges} in the graph to be colored that are not listed in the prompt, which can lead them to mistake a colorable graph for an uncolorable one. (To use the terminology of \citet{zhang2025siren}, these are \textit{input-conflicting hallucinations}, reflecting a failure of models to properly deploy information they are explicitly given at runtime, which are distinct from the better-studied \textit{fact-conflicting hallucinations} that reflect failure to properly memorize information during training.) This phenomenon persists across multiple problem complexity levels and multiple semantic frames in which graph coloring problems can be presented, and it appears to account for at least a significant fraction of the incorrect answers given by every model, and the vast majority of incorrect answers for some models. We present our findings across a series of experiments (including smaller-scale experiments that demonstrate a similar hallucination phenomenon in a type of stable matching problem; see Appendix \ref{app:stable_matching}) and conclude by discussing our results' broader implications for RLLM reasoning and how its flaws might be addressed.

\section{General Methodology} \label{sec:general}

We conduct our experiments on the following RLLMs: (1) OpenAI's o1-mini \citep{openai2024o1mini}, specifically \texttt{o1-mini-2024-09-12}, through the OpenAI API; (2) OpenAI's o3-mini \citep{openai2025o3mini}, specifically \texttt{o3-mini-2025-01-31}, through the OpenAI API; (3) DeepSeek-R1 \citep{deepseek2025r1} through a combination of the Fireworks AI and DeepInfra online hosts; (4) Anthropic's Claude 3.7 Sonnet \citep{anthropic2025claude3.7s}, specifically \texttt{claude-3-7-sonnet-20250219} with thinking mode turned on, through the Anthropic API; (5) Google's Gemini 2.5 Pro Preview \citep{google2025gemini2.5} through the Gemini Developer API; and (6) xAI's Grok 3 Mini Beta \citep{xai2025grok3beta} through the xAI API. For o3-mini and Grok 3 Mini Beta, we test all reasoning effort settings (low, medium, and high for the former, and low and high for the latter). We use the mini counterparts of o1, o3, and Grok 3 Beta rather than the full-size models for cost-effectiveness reasons and because the mini models are specialized for logical and mathematical reasoning; we do not expect the broader knowledge of the full-size models to be particularly helpful for the graph coloring task.

As in \citet{heyman2025evaluating}, for models with variable temperature settings, we set temperature to 0 to match the low-creativity deductive nature of the graph coloring task, with the exception of DeepSeek-R1; its developers recommend a temperature no lower than 0.5 \citep{deepseek2025r1modelcard}, and we use a temperature of 0.5 for this reason. We find non-terminating output from our prompts at these temperatures to be infrequent for DeepSeek-R1 and nonexistent for all other models.

Claude 3.7 Sonnet requires the user to specify a maximum response length and, when thinking mode is turned on, a target thinking budget. We set these values to 3000 and 2000 tokens respectively for 4-vertex problems, and to 9000 and 8000 tokens for 8-vertex problems. Of the 6600 8-vertex responses we collected in total, 421 were cut off by the length limit; these were re-run with the parameters doubled to 18000 and 16000 tokens, and none were cut off then.

We use the same prompting format as in \citet{heyman2025evaluating}; the model is given a description of a graph, including a fully enumerated and clearly labeled set of edges, then asked if it is possible to color the graph with a specified set of colors so that no two adjacent vertices receive the same color, and to give a plan for it if so. To broaden our observations, we deploy prompts separately with two different frames, ``Math" and ``Friends", which we designed to probe LLM semantic biases. ``Math" prompts present the problem explicitly as coloring a graph with edges connecting numbered vertices; meanwhile, in deliberately less-orthodox ``Friends" prompts, the model is given a list of friendships that exist between $n$ people (named alphabetically as Alice, Bob, Carol, etc.) and told that everyone is going to wear a shirt with one of $k$ colors and no one wants to wear the same color shirt as any of their friends. (Examples of the prompts we use can be found in Appendix \ref{app:example_prompts}.) When evaluating model responses for answer correctness, we use an automatic parser and manually review and interpret any responses in which it does not detect a coherent answer.

All error bars in our figures represent $95\%$ Clopper-Pearson binomial confidence intervals with the number of trials $n$ equal to the total number of prompt submissions and corresponding model responses in the category plotted.

Code for reproducing our experiments and records of our problems, prompts, and responses (including those in appendices) can be found at \url{https://github.com/AlexHeyman/RLLMGraphColoring}. The vast majority of the compute used in our experiments was on the model API side, making estimation of its magnitude difficult. However, each experiment section specifies the number of prompt submissions to the RLLMs involved; the totals across all experiments are 5,155 for o1-mini, 7,155 for DeepSeek-R1, 7,651 for Claude 3.7 S, 9,230 for Grok 3 MB (low), 14,230 for Grok 3 MB (high), 12,230 for o3-mini (high), 7,230 for each of o3-mini (low), o3-mini (medium), and Gemini 2.5 PP, and 2,000 for each of gpt-oss-120b (low), (medium), and (high) in Appendix \ref{app:stable_matching}. All of our data analysis takes no more than 1 hour to run on a single AMD Ryzen 9 8945HS CPU.

\section{Experiment 1: Basic Error Patterns}

\subsection{Purpose}

In \citet{heyman2025evaluating}, we find several basic patterns in the errors made by both o1-mini and DeepSeek-R1, including (1) errors on 4-vertex 2-coloring problems are very rare but not impossible; (2) on 8-vertex 4-coloring problems, errors are much more frequent and mostly consist of declaring colorable problems to be uncolorable; and (3) error rates are generally higher for the Friends frame than the Math frame (which we hypothesize results from the greater prevalence of explicitly mathematical reasoning problems in the models' training). To start, we want to see if these patterns hold for the other RLLMs we test in this work, and how their error rates compare to o1-mini and DeepSeek-R1.

\subsection{Methods}

We test Claude 3.7 Sonnet, o3-mini (low, medium, and high), Gemini 2.5 Pro Preview, and Grok 3 Mini Beta (low and high) on the 4-vertex 2-coloring (``4v2c") and 8-vertex 4-coloring (``8v4c") problem sets on which we also tested o1-mini and DeepSeek-R1 in \citet{heyman2025evaluating}. The 4v2c problem set consists of all possible 4-vertex graphs counting vertex-order permutations separately, minus the graph with no edges, for a total of $2^{C(4, 2)} - 1 = 63$ problems (40 colorable and 23 uncolorable). The version of the 8v4c problem set on which we test our RLLMs consists of random samples of 50 8-vertex graphs from each possible edge count between 14 and 23 inclusive (out of a maximum of $C(8, 2) = 28$ edges), for a total of 500 problems (306 colorable and 194 uncolorable). We prompt each model with each problem in each frame five separate times to increase statistical power given model stochasticity. We then evaluate the models' responses and calculate the relevant error rates. We include the responses from o1-mini and DeepSeek-R1 that we analyzed in \citet{heyman2025evaluating} in our analysis here as well.

\subsection{Results}

\begin{figure}[t]
\begin{centering}
\includegraphics[width=\textwidth]{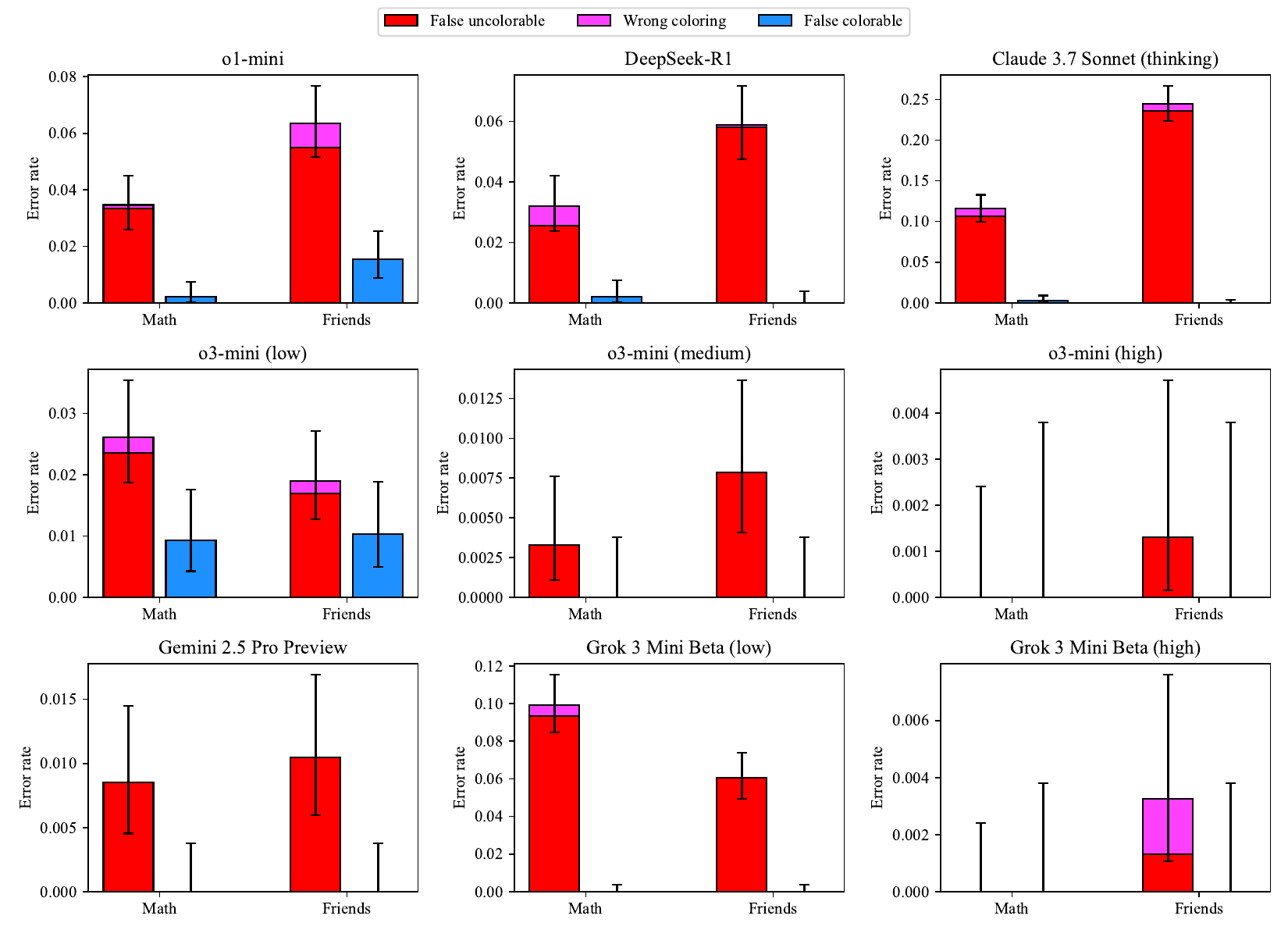}
\caption{Rate of each error type on the 8v4c problem set for each model and frame. Error types are grouped by whether they apply to colorable or uncolorable problems (left and right of each pair of bars, respectively), and error rates are calculated as a fraction of the trials on the corresponding problem type. Note that each subplot's y-axis is independently scaled to its bar heights.}
\label{fig:8v4c_error_types}
\end{centering}
\vspace{-0.1in}
\end{figure}

We found that, like with o1-mini and DeepSeek-R1, the newer RLLMs yielded error rates no greater than $1$--$2\%$ on the simple 4v2c problem set; indeed, most of them surpassed the older two and gave no incorrect answers at all, with the exception of o3-mini (low) and (medium). A full table of error counts by model, frame, and error type can be found in Appendix \ref{app:errors_on_4v2c}, though most cell entries are 0.

Rates of different error types for each model and frame on 8v4c are plotted in Figure \ref{fig:8v4c_error_types} (note the varying y-axis scales). Errors are categorized into \textit{false colorable} (giving a coloring for an uncolorable problem), \textit{wrong coloring} (giving an invalid coloring for a problem that has valid colorings), and \textit{false uncolorable} (declaring a colorable problem to be uncolorable). Error rates vary widely between models here, with Claude 3.7 S reaching $>10\%$ error on Math colorable problems and $>20\%$ error on Friends colorable problems, while o3-mini (high) and Grok 3 MB (high) each make no errors at all on Math, and only 2 and 5 errors on Friends respectively (out of 2500 total trials and 1530 colorable trials). However, the pattern of most errors (both in absolute terms and relative to applicable trials) being false-uncolorable reappears in almost all of the newer RLLMs alongside o1-mini and DeepSeek-R1; the only exception is Grok 3 MB (high), with an error sample size of only 5 (2 false uncolorable and 3 wrong coloring). Not surprisingly, both o3-mini and Grok 3 MB exhibit drastically decreasing error rates as reasoning effort increases. In terms of the relationship between frames, o1-mini and DeepSeek-R1's pattern of more frequent errors on Friends than on Math is shared by Claude 3.7 Sonnet and (possibly, considering the confidence intervals) Gemini 2.5 PP, o3-mini (medium and high), and Grok 3 MB (high); however, Grok 3 MB (low) and (possibly) o3-mini (low) exhibit lower error on Friends instead.

Note that, for Gemini 2.5 PP and Grok 3 MB, some of the repeated trials for the same problem and frame yielded the exact same response word-for-word; for instance, for Grok 3 MB (high) on 8v4c Friends, the two false-uncolorable errors come from two identical responses to the same problem, and the same is true of the three wrong-coloring errors. Neither of these models' API documentation mentions any full-response-caching behavior; the repetition may be because we ran both models at temperature 0. We treat these repetitions as independent trials because what we are interested in is the \textit{probability} of a model's output having certain features given that its prompt had certain features.

\section{Experiment 2: Edge Hallucinations}

\subsection{Purpose}

We conducted informal analysis of the text of the erroneous DeepSeek-R1 responses included in Experiment 1, and found that on 8v4c problems, the model's chain of thought sometimes claims that the graph contains a particular edge that the prompt does not list. Such additional edges would create additional constraints on valid colorings for the graph, and indeed these ``edge hallucinations" can lead DeepSeek-R1 to conclude that problems are uncolorable when they are actually colorable. We want to know how common edge hallucinations are in the responses of DeepSeek-R1 and the other RLLMs we test, and how many of the RLLMs' false-uncolorable errors can be attributed to them. If edge hallucinations are common (more common than models ignoring edges that do exist), it could be at least part of the explanation for why false-uncolorable errors are the most common type.

\subsection{Methods}

We search for edge hallucinations in the 8v4c problem responses included in Experiment 1. Because of the high time cost of manually screening model response text for erroneous statements, we search using a combination of manual review and hand-coded parsers that scan response text for claims about the existence of certain edges. Apparent false claims found by the parsers are manually reviewed for false positives and the parsers are also iteratively tuned on this basis, meaning that our estimates of edge hallucination rates function as approximate lower bounds. Some of the RLLMs we test -- o1-mini, o3-mini, and Gemini 2.5 Pro Preview -- do not expose their chains of thought, meaning that edge hallucination searches must rely on the explanations these models give alongside their final answers. Of these models, only o3-mini (high) and Gemini 2.5 PP reliably give answer explanations detailed enough to check for edge hallucinations in, and because these explanations are less consistently structured than chains of thought, they are less amenable to automatic parsing. Thus, we do not attempt to gather edge hallucination data for o1-mini, o3-mini (low), and o3-mini (medium), and we only gather data for responses with false-uncolorable final answers for o3-mini (high) and Gemini 2.5 PP. Additionally, because detecting the \textit{absence} of an edge in a model's reasoning process from reading its response text is harder than detecting the \textit{presence} of one, we do not attempt to search for ``edge forgetting" -- though, as we will see, indirect evidence indicates it is much less common than the hallucination of nonexistent edges.

\subsection{Results}

\begin{figure}[t]
\begin{centering}
\includegraphics[width=\textwidth]{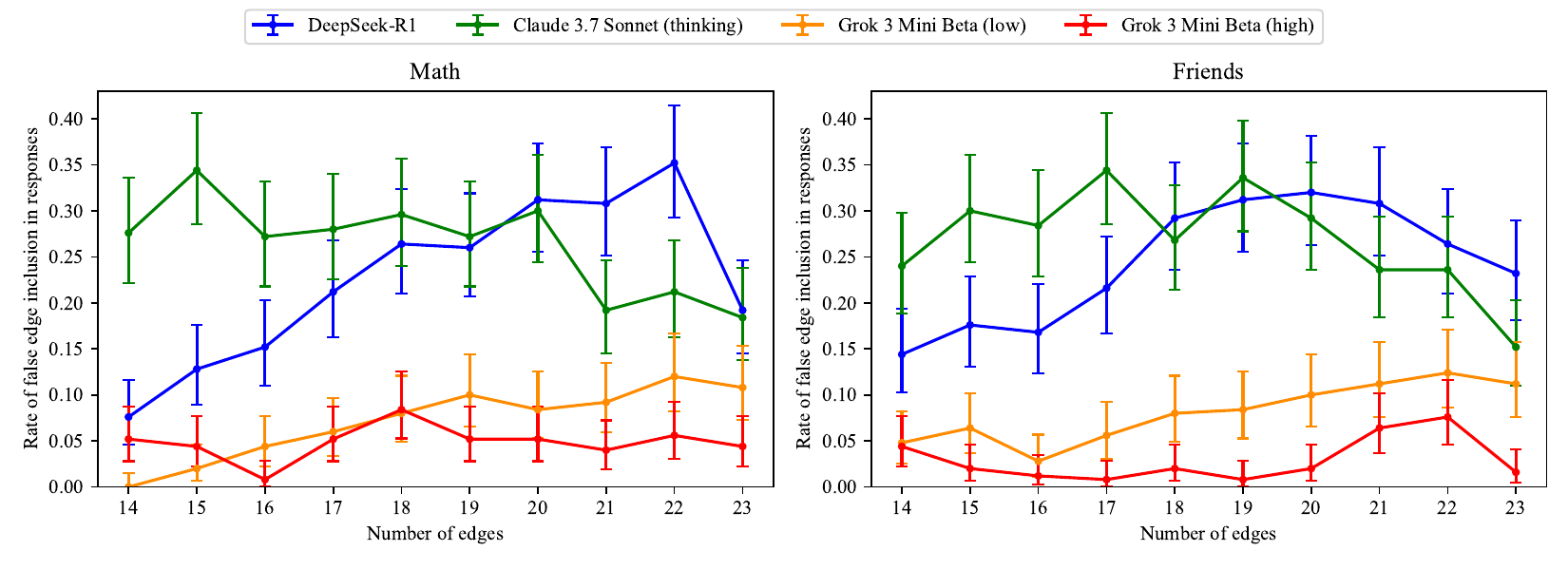}
\caption{Rates of responses containing edge hallucinations by edge count in the 8v4c problem set, for each frame and model with exposed CoT.}
\label{fig:8v4c_hallucinations}
\end{centering}
\vspace{-0.1in}
\end{figure}

Plots of 8v4c edge hallucination rates (proportions of responses claiming the existence of at least one false edge) against problems' true number of edges, for each frame and CoT-exposing model, can be seen in Figure \ref{fig:8v4c_hallucinations}. Models differ greatly in overall edge hallucination rates, but all of them hallucinate edges in at least some cases. The relationship between hallucination rate and edge count varies by model and frame -- we might expect that hallucinations become more common the more real edges the model has to keep track of, and we might also expect that they become less common the fewer possible but nonexistent edges there \textit{are} to hallucinate; the balance of these factors and/or other unknown ones may differ between models and frames.

\begin{figure}[t]
\begin{centering}
\includegraphics[width=\textwidth]{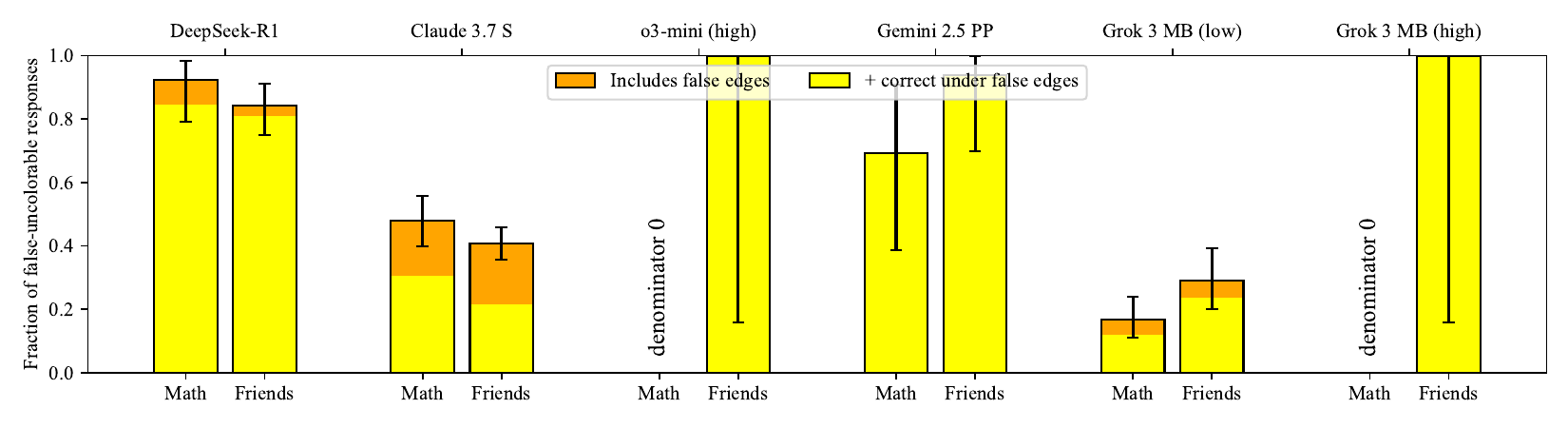}
\caption{Attributability of false-uncolorable errors to edge hallucinations in the 8v4c problem set, for each frame and model with data on false edges for false-uncolorable responses.}
\label{fig:8v4c_attributability}
\end{centering}
\vspace{-0.1in}
\end{figure}

Figure \ref{fig:8v4c_attributability} shows, for each frame and model with false-uncolorable false edge data, the proportion of false-uncolorable responses that include edge hallucinations, and the proportion that furthermore would answer correctly if the hallucinated edge(s) actually existed (i.e. the graph really would be uncolorable then). A response falling into the ``would be correct" category is evidence (though not proof) that its edge hallucinations are the \textit{cause} of its incorrectness. Depending on the model and frame, the ``would be correct" proportion of false-uncolorable responses can be as many as 33 out of 39 (DeepSeek-R1 Math) or 15 out of 16 (Gemini 2.5 PP Friends), or as low as 17 out of 143 (Grok 3 MB (low) Math) -- but it is at least significantly above 0 in all cases with data available. Notably, for all combinations of frame and model with exposed CoT, there were many responses that included edge hallucinations but did not give false-uncolorable answers; causes of this can include the graph being uncolorable to begin with, the added edges still leaving the graph colorable, or the model's chain of thought recognizing and correcting its mistake before completing (which we have informally observed in a considerable number of responses, but which would be difficult to systematically screen for).

Also of note is that, for every model-frame combination with the relevant data available, the distribution of \textit{which edges} get hallucinated is distinctly non-uniform, even though the problem graphs were generated in a completely vertex-symmetric manner and thus the distribution of which edges do not exist \textit{is} roughly uniform. The tendency is for edges involving the vertices with the \textit{highest indices} to get hallucinated most; in the Math frame, with vertices numbered from 0-7 inclusive, the edge that a given model most frequently hallucinates is generally (6, 7), with (5, 6) and to a lesser extent (5, 7) also popular. Similarly, in the Friends frame, the most popular edge to hallucinate is (George, Heather), followed by (Fran, George) and (Fran, Heather). Confirming the causal significance of this striking cross-model pattern is part of the goal of the next experiment we recount.

\section{Experiment 3: Problems Targeting Edge Hallucination}

\subsection{Purpose}

Because chains of thought are not guaranteed to accurately report model decision-making strategies (\citet{chen2025reasoning}, as discussed previously) and post-CoT explanations are generated after the model is supposed to have determined its answer, we want to verify the causal significance of the edge hallucination patterns we have identified using a method that does not rely on analyzing response text beyond final answers. We pursue this by constructing problem sets designed to increase the likelihood that edge hallucination will result in an incorrect answer, and seeing if and to what extent error rates indeed increase on these problem sets.

\subsection{Methods}

We generate two additional sets of 8-vertex 4-coloring problems. The first, called ``8v4c high-edge-count colorable" (8v4c HECC), consists of 100 8-vertex graphs from each edge count between 20 and 23 inclusive, generated randomly but rejection-sampled to ensure that they are colorable. The second, called ``8v4c adversarial", is generated identically to 8v4c HECC, but with an additional rejection sampling condition: the graph must not contain the edge (6, 7), and it must be uncolorable when (6, 7) is added. We computationally verify that for every graph in 8v4c HECC, there is at least one edge that would make the graph uncolorable if added; however, this is not the case for most colorable problems with lower edge counts in the base 8v4c problem set. This means that if edge hallucination can cause false-uncolorable errors in the models we test, it should be especially likely to cause them in 8v4c HECC as opposed to colorable problems in base 8v4c, and more likely still in 8v4c adversarial if the hallucinated-edge identity bias discussed in Experiment 2's results holds.

We test all of our models on these problem sets, including o1-mini and DeepSeek-R1. We repeat the tests in both frames, but prompt with each problem-frame combination only once rather than five times; we judge that the additional statistical power is not worth the additional financial and computational cost in this case. We also search for edge hallucinations in the text of false-uncolorable responses from models where it is feasible, using the same methodology as in Experiment 2.

\subsection{Results}

\begin{figure}[t]
\begin{centering}
\includegraphics[width=\textwidth]{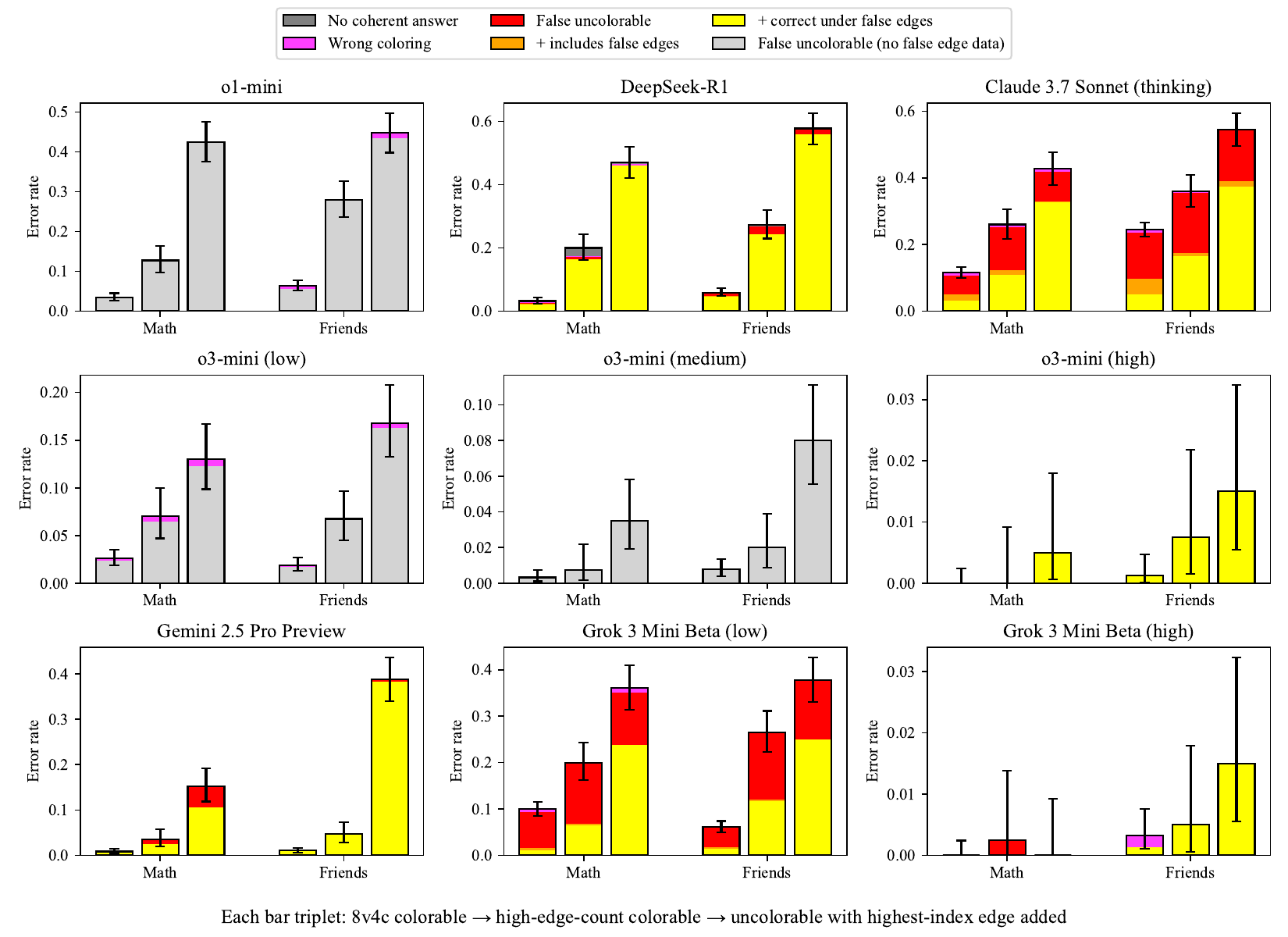}
\caption{Error rates by type and false-uncolorable attributability for colorable problems in the base 8v4c, 8v4c high-edge-count colorable, and 8v4c adversarial problem sets, for each model and frame. Note that each subplot's y-axis is independently scaled to its bar heights.}
\label{fig:8v4c_progression}
\end{centering}
\vspace{-0.1in}
\end{figure}

Figure \ref{fig:8v4c_progression} shows plots, for each model and frame, of error rates and false-uncolorable attributability for colorable problems in base 8v4c, 8v4c HECC, and 8v4c adversarial (left, center, and right of each bar triplet, respectively). Across almost all models and frames, rates of both overall error and false-uncolorable error (which are usually almost the same) distinctly increase from base 8v4c to 8v4c HECC and again to 8v4c adversarial. This pattern is less clear in model-frame combinations with fewer errors overall, but given its overall consistency across models and frames, this is likely just an artifact of sample size. Furthermore, the proportion of false-uncolorable responses found to contain edge hallucinations in their text that would make them correct is high for the new problem sets wherever it is high for base 8v4c -- and for Claude 3.7 S and Grok 3 MB (low), where this proportion is low for base 8v4c, it increases greatly from base to HECC to adversarial. This suggests that these models frequently make false-uncolorable errors that are \textit{not} due to edge hallucinations, but the edge-hallucination-based errors come to dominate in problem sets designed to invite them.

\section{Experiment 4: Higher Problem Complexity}

\subsection{Purpose}

We want to see if and how the error-type distribution and edge hallucination phenomena we have found generalize to problem complexities higher than 8 vertices and 4 colors. We also want to gather more data on errors made by o3-mini (high) and Grok 3 MB (high), the two models for which the cross-model patterns we have observed have been least clear due to low error sample sizes. To this end, we generate a new, higher-complexity graph coloring problem set and test these models on it.

\subsection{Methods}

We generate a set of problems with 12 vertices and 6 available colors (``12v6c"). Analogously to 8v4c, it consists of random samples of 50 12-vertex graphs from each edge count between 47 and 56 inclusive (out of a maximum of $C(12, 2) = 66$ edges), for a total of 500 problems (coincidentally, exactly 250 colorable and 250 uncolorable). We prompt each of o3-mini (high) and Grok 3 MB (high) with each combination of problem and frame five times, and search the text of false-uncolorable responses for edge hallucinations using the same methodology as in previous experiments.

\subsection{Results}

\begin{figure}[t]
\begin{centering}
\includegraphics[width=\textwidth]{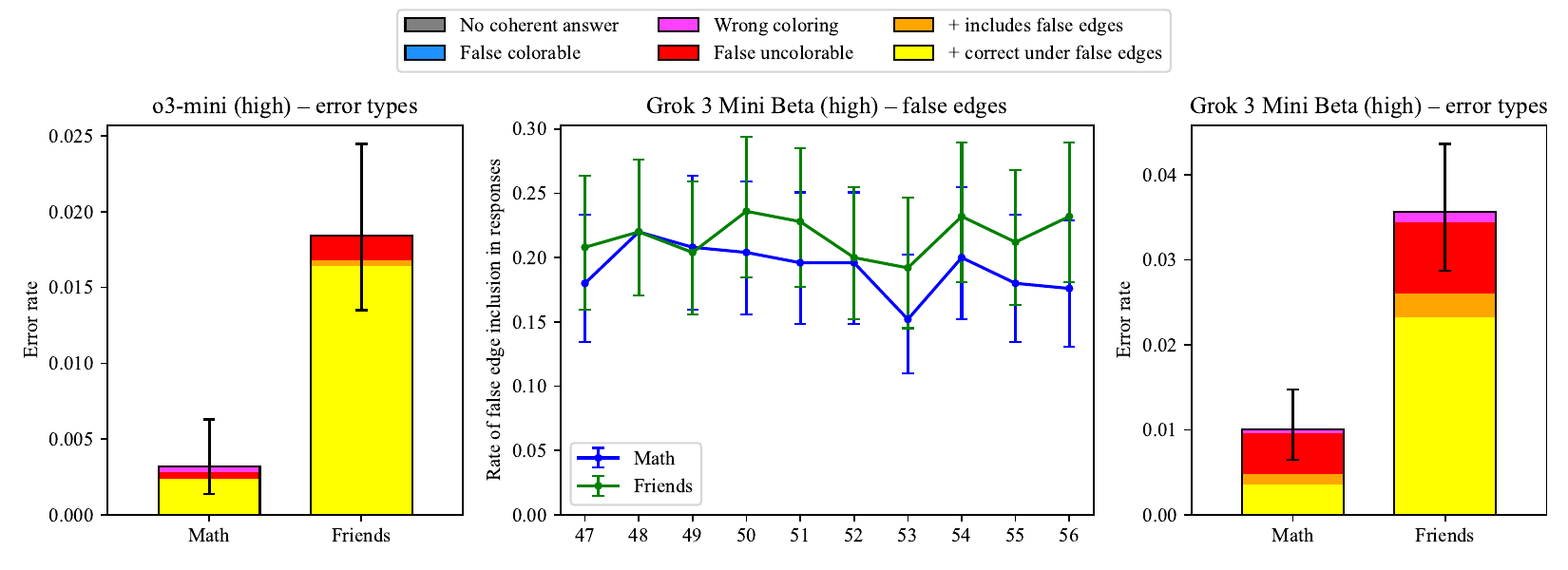}
\caption{\textbf{Left:} Error rates by type and false-uncolorable attributability for o3-mini (high) on 12v6c. \textbf{Center:} Edge hallucination rate by edge count for Grok 3 MB (high) on 12v6c. \textbf{Right:} Error rates by type and false-uncolorable attributability for Grok 3 MB (high) on 12v6c.}
\label{fig:12v6c}
\end{centering}
\vspace{-0.1in}
\end{figure}

Plots of error rates by type and false-uncolorable attributability for both models on 12v6c, and of edge hallucination rate for Grok 3 MB (high) thanks to its exposed CoT, can be seen in Figure \ref{fig:12v6c}. Overall error rates are no more than a few percent for any model-frame combination, but they are much higher than these models' error rates on the simpler 8v4c problem set. As with models in general on 8v4c, errors are dominated by the false-uncolorable type. Rates of detected edge hallucinations for Grok 3 MB (high) are considerably higher for both frames on 12v6c than on 8v4c, and most false-uncolorable responses contain edge hallucinations that would make them correct (except for Grok 3 MB (high) Math, where the rate is 9 out of 24 responses). For all four model-frame combinations, the most frequently hallucinated edge is (10, 11), or the equivalent (Kathy, Larry) for Friends; the rest of the frequency distribution varies, but as with 8v4c, the distribution is always distinctly non-uniform.

\section{Discussion}

We have shown that a variety of RLLMs from different developers and with release dates spanning the majority of RLLM history are prone to hallucinate edges in (sufficiently complex) graph coloring problems, which can lead them to answer the problems incorrectly. This phenomenon occurs even if the problems are not stated explicitly in terms of graph coloring, and it persists and appears to even grow more frequent as problem complexity increases. The presence of such a straightforward input-conflicting hallucination phenomenon in LLMs specialized for solving reasoning problems is striking, and given the lengths of the prompts and responses it cannot be explained away by context window limitations. Instead, it may indicate that RLLMs have issues distinguishing appropriately between information in the prompt and the potentially less reliable information in their own in-progress chain of thought; if so, they may benefit from a feature akin to positional encoding that explicitly demarcates prompt tokens vs. chain-of-thought tokens, if they do not already have it. The asymmetry in which edges they tend to hallucinate may also reflect weaknesses in how they represent numbers or orderings, with higher-indexed entities becoming more easily conflated with one another.

We also know that the hallucination phenomenon we identify is not exclusive to graph coloring, owing to the supplementary experiments we describe in Appendix \ref{app:stable_matching}. We generate stable matching problems in which students must be assigned to dorm rooms in groups of 3, and find that a significant source of RLLM error on them is misrepresentation of students' ratings for each other -- most often transposing \textit{A}'s rating for \textit{B} onto \textit{B}'s rating for \textit{A}, suggesting weaknesses in representing directionality.

\subsection{Limitations}

We do not test all publicly available RLLMs, and we only test two problem types (graph coloring and stable 3-matching) as representatives of the category of logic tasks. However, the existence and behavior of the hallucinations we observe are consistent enough in the RLLMs we do test that we expect them to occur in any other similarly designed RLLMs, and we find it unlikely that the design flaws that enable these hallucinations do not also affect at least some other problem types on which the RLLMs have received comparably thorough or less thorough training.

\subsection{Societal Impacts}

Research identifying the limitations of RLLMs is valuable for discouraging dangerous overreliance on them; however, the application of this research for improving RLLM capabilities may amplify the technology's social and economic effects, including negative ones. Responsibility for mitigating these harms lies with governments and with the AI research and development community.

\begin{ack}
This work was supported by a Canada Research Chair Grant (CRC-2022-00277) to JZ. JZ also gratefully acknowledges the support of CIFAR, in his role as CIFAR Fellow of Learning in Machines and Brains.
\end{ack}

\bibliography{paper}
\bibliographystyle{plainnat}


\newpage
\appendix

\section{Experiment 0: Greedy Colorability and Error Rate} \label{app:greedy_score}

\begin{figure}[h]
\begin{centering}
\includegraphics[width=\textwidth]{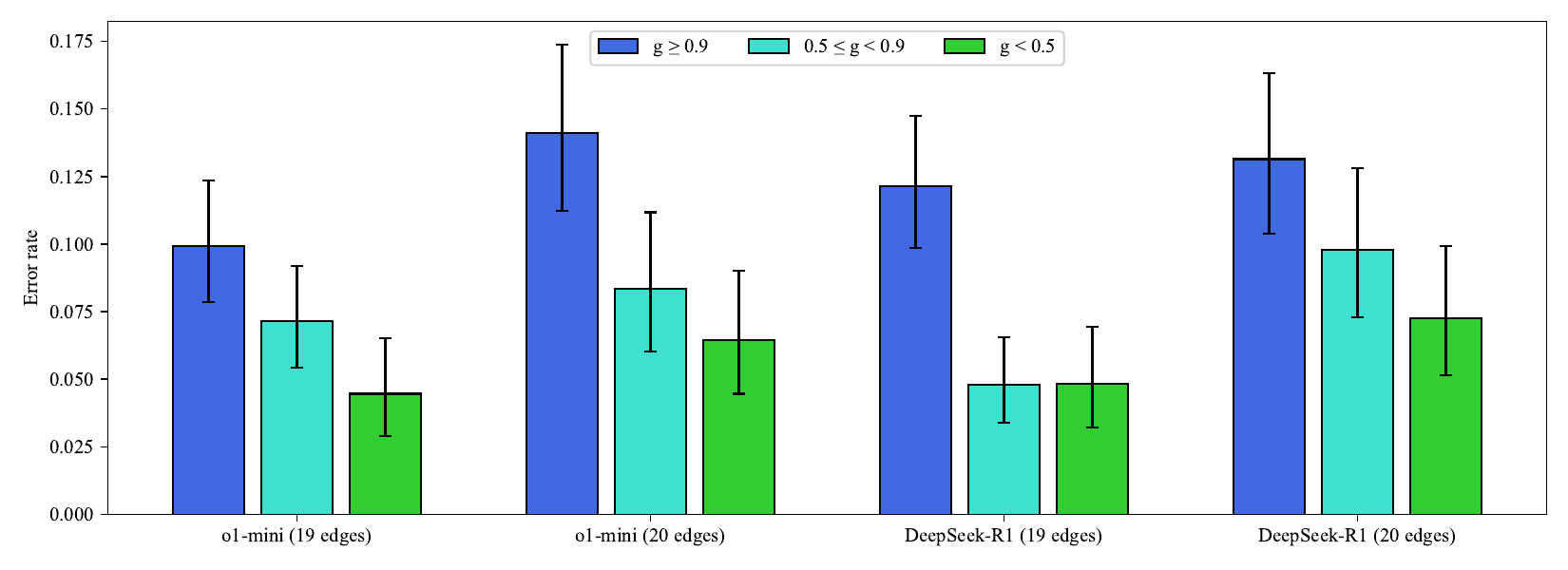}
\caption{Error rate for each greedy score category and edge count in our validation problem set, for o1-mini and DeepSeek-R1 each.}
\label{fig:greedy_score_test}
\end{centering}
\vspace{-0.1in}
\end{figure}

\subsection{Purpose}

In \citet{heyman2025evaluating}, we categorize colorable graph coloring problems using a numerical estimate of how easy they are to successfully color ``greedily" one vertex at a time without ever backtracking on coloring choices, naming this measure the ``greedy score". We find that among colorable problems, both o1-mini and DeepSeek-R1's error rates increase drastically as greedy score decreases (i.e. as greedy coloring becomes harder), more sharply than with the traditional LLMs we test. We initially took this as evidence that though the two RLLMs represent a large improvement in step-by-step reasoning ability, their gains in possibility space exploration have not been as great. However, the correlation we observed is across problems with varying edge counts, and high edge count also correlates positively with error among colorable problems. Here, we want to see if low greedy score remains related to high error when edge count is controlled for.

\subsection{Methods}

We construct a validation problem set by randomly sampling 1,000 8-vertex graphs with 19 edges and 1,000 with 20 edges, for a total of 2,000, for use in 4-coloring problems. We then calculate the greedy score $g$ for all of the colorable graphs and divide them into the same three categories we use in \citet{heyman2025evaluating}: $g \geq 0.9$, $0.5 \leq g < 0.9$, and $g < 0.5$, containing 250, 676, and 211 problems respectively. We randomly sample 250 of the 676 problems from the middle category to even out the category sizes, leaving a total of 711 problems (19 edges: 145, 154, and 112 for the three categories; 20 edges: 105, 96, and 99). We prompt each of o1-mini and DeepSeek-R1 with each problem five times each, using the same methodology as in Section \ref{sec:general}, and only in the Math frame.

\subsection{Results}

Plots of error rates for each combination of model, edge count, and greedy score category can be seen in Figure \ref{fig:greedy_score_test}. Contrary to our earlier hypothesis, for every combination of model and edge count, error rate tends to be \textit{lower} on problems with lower greedy score, which should be \textit{harder} to find valid colorings for if the models' primary area of weakness was systematic possibility space exploration in the form of backtracking or other non-greedy strategies. Less surprisingly, however, error rate is always at least slightly higher for 20-edge problems than 19-edge problems when model and greedy score category are held constant. We do not know why greedy score and error rate distinctly correlate in the opposite-of-expected direction, rather than not distinctly correlating at all. We also cannot \textit{guarantee} that this pattern holds for other models, frames, or edge/vertex/color counts with a diversity of greedy scores, but neither do we have any particular reason to expect this experimental setup to be unrepresentative.

\newpage
\section{Example Prompts} \label{app:example_prompts}

Following are two examples of the prompts we give to the models we test, representing a particular problem from the 8v4c problem set in the Math and Friends frames each.

\subsection{Math}

\noindent\fbox{
\parbox{\textwidth}{
\begin{allintypewriter}
Consider an undirected graph with 8 vertices (numbered 0 through 7) and the following set of edges:\\

\{(0,3), (0,5), (0,6), (0,7), (1,2), (1,3), (1,4), (1,5), (1,7), (2,3), (2,5), (2,6), (2,7), (3,4), (3,5), (3,7), (4,5), (4,6), (4,7), (5,6), (6,7)\}\\

Suppose that we want to color every vertex either red, green, blue, or yellow so that no two adjacent vertices receive the same color. Is this possible? If it is impossible, write "Impossible" as the final line of your response. If it is possible, the final lines of your response should present a plan for it in a format like the following:\\

0 Red\\
1 Green\\
2 Blue\\
3 Yellow\\
(etc.)
\end{allintypewriter}
}
}

\subsection{Friends}

\noindent\fbox{
\parbox{\textwidth}{
\begin{allintypewriter}
Imagine 8 people: Alice, Bob, Carol, Dave, Ethan, Fran, George, and Heather. Suppose that the following friendships exist between them: Alice is friends with Dave, Alice is friends with Fran, Alice is friends with George, Alice is friends with Heather, Bob is friends with Carol, Bob is friends with Dave, Bob is friends with Ethan, Bob is friends with Fran, Bob is friends with Heather, Carol is friends with Dave, Carol is friends with Fran, Carol is friends with George, Carol is friends with Heather, Dave is friends with Ethan, Dave is friends with Fran, Dave is friends with Heather, Ethan is friends with Fran, Ethan is friends with George, Ethan is friends with Heather, Fran is friends with George, and George is friends with Heather.\\

Suppose that the 8 people are all going to attend a party, and each of them is going to wear either a red shirt, a green shirt, a blue shirt, or a yellow shirt. Suppose that none of them want to wear the same color shirt as anyone they are friends with. Is this possible? If it is impossible, write "Impossible" as the final line of your response. If it is possible, the final lines of your response should present a plan for it in a format like the following:\\

Alice: Red\\
Bob: Green\\
Carol: Blue\\
Dave: Yellow\\
(etc.)
\end{allintypewriter}
}
}

\newpage
\section{Errors on 4v2c} \label{app:errors_on_4v2c}

Following is a table of absolute counts of errors for each model and frame on the 4v2c problem set, both total errors and specific error types. The total number of trials for each model-frame combination is $63 \times 5 = 315$ (200 trials on colorable problems and 115 on uncolorable). In the column labels, ``f.c." refers to false colorable errors, ``w.c." to wrong coloring, and ``f.u." to false uncolorable.

\begin{table}[h]
\caption{Error counts on 4v2c}
\label{tab:4v2c}
\centering
\begin{tabular}{>{\centering\arraybackslash}p{0.18\textwidth} >{\centering\arraybackslash}p{0.065\textwidth} >{\centering\arraybackslash}p{0.065\textwidth} >{\centering\arraybackslash}p{0.065\textwidth} >{\centering\arraybackslash}p{0.065\textwidth} >{\centering\arraybackslash}p{0.065\textwidth} >{\centering\arraybackslash}p{0.065\textwidth} >{\centering\arraybackslash}p{0.065\textwidth} >{\centering\arraybackslash}p{0.065\textwidth}}
\hline
Model & Math (total) & Math (f.c.) & Math (w.c.) & Math (f.u.) & Friends (total) & Friends (f.c.) & Friends (w.c.) & Friends (f.u.)\\
\hline
o1-mini & 3 & 1 & 2 & 0 & 4 & 1 & 3 & 0\\
DeepSeek-R1 & 2 & 0 & 2 & 0 & 1 & 0 & 1 & 0 \\
Claude 3.7 S & 0 & 0 & 0 & 0 & 0 & 0 & 0 & 0 \\
o3-mini (low) & 5 & 5 & 0 & 0 & 2 & 2 & 0 & 0 \\
o3-mini (med) & 1 & 1 & 0 & 0 & 0 & 0 & 0 & 0 \\
o3-mini (high) & 0 & 0 & 0 & 0 & 0 & 0 & 0 & 0 \\
Gemini 2.5 PP & 0 & 0 & 0 & 0 & 0 & 0 & 0 & 0 \\
Grok 3 MB (low) & 0 & 0 & 0 & 0 & 0 & 0 & 0 & 0 \\
Grok 3 MB (high) & 0 & 0 & 0 & 0 & 0 & 0 & 0 & 0 \\
\hline
\end{tabular}
\end{table}

\section{Rating Hallucinations in Stable 3-Matching Problems} \label{app:stable_matching}

After completing the experiments on graph coloring problems that we describe in this work's main body, we took interest in using similar methodology to analyze RLLM behavior on a different type of reasoning problem, to test our hypothesis that phenomena akin to edge hallucination generalize beyond graph coloring and to potentially gain a broader perspective on RLLM reasoning from the effects (or lack thereof) on RLLM behavior of the differences between graph coloring and the new problem type.

The new problem type we chose is a stable matching problem in which objects must be assembled into groups of 3, rather than groups of 2 as is typical for such problems. Stable 3-matching problems have been studied in a theoretical computer science context (see e.g. \citet{ng1991three}), but to our knowledge, no previous work has discussed the specific form we use here. We frame our problem type in terms of $3n$ university students who must be grouped into $n$ dorm rooms holding 3 students each. Each student has rated each other student on a scale of 1 to 5 (note that a student can rate multiple other students equally, and two students' ratings for each other need not be equal), and a student's satisfaction with their room is the sum of their ratings for their two roommates. A grouping of students into rooms is unstable if and only if there exists a pair of students who could switch rooms to increase both of their satisfactions, or increase one and keep the other the same. The goal of the problem is to give an example of a stable grouping if one exists, or to report that no stable grouping exists. Because this specific problem type has not been a subject of past research, we expect RLLMs to be unfamiliar with it from their training compared to the basic form of graph coloring we have tested them on, which has been studied and discussed extensively elsewhere; we are interested in observing potential differences in RLLM behavior arising from this lower familiarity. Also of note is that we know of no algorithm for solving this problem type without fail that does not in the worst case involve checking every possible grouping, though worthwhile heuristics exist.

We generate two stable 3-matching problem sets of differing complexity and test several RLLMs with exposed chains of thought on them. We find that, similar to edge hallucination in graph coloring, the RLLMs' chains of thought sometimes contain false claims about students' ratings, which are most often transpositions of some student \textit{A}'s rating for another student \textit{B} onto \textit{B}'s rating for \textit{A}. These rating hallucinations can cause RLLMs to give incorrect final answers, though when the rate of incorrect final answers is high, it tends to be the case that few of them can be attributed \textit{solely} to rating hallucinations, indicating major influence from other types of reasoning mistakes. These results support our hypothesis that input-conflicting hallucinations are a significant source of RLLM error across a broad range of reasoning problems, while also adding nuance to the evidence base on the role of other error sources.

\subsection{Problem Sets}

We procedurally generate two problem sets of 1,000 problems each: a lower-complexity one in which 6 students must be assembled into 2 groups of 3, and a higher-complexity one in which 9 students must be assembled into 3 groups of 3. We respectively call these problem sets ``2g3" and ``3g3" for short. To generate each problem, each student's rating for each other student is sampled independently and uniformly at random from the integers between 1 and 5 inclusive.

In each 2g3 problem, there are $6 \times (6 - 1) = 30$ different rating values for the model to distinguish between and $\frac{6!}{(3!)^2 \times 2!} = 10$ possible groupings of students to check in the worst case. Of the 1,000 2g3 problems we generated, 288 are such that none of the 10 groupings are stable, 447 have 1 stable grouping, 210 have 2 stable groupings, and 55 have 3 or more.

In each 3g3 problem, there are $9 \times (9 - 1) = 72$ different rating values to distinguish between and $\frac{9!}{(3!)^3 \times 3!} = 280$ possible groupings to check in the worst case. Because of the large number of possible groupings, we would not expect even an ``ideal" language model of a presently-realistic size to be able to solve 3g3 problems universally without fail, at least not without a means of bypassing its context window limits; however, we are still interested in studying exactly how and how much real RLLMs err on this problem set. Of the 1,000 3g3 problems we generated, 384 have 0 stable groupings out of 280, 326 have 1 stable grouping, 160 have 2, 68 have 3, and 62 have 4 or more.

\subsection{Models}

We test DeepSeek-R1 (through the DeepInfra online host) and Grok 3 Mini Beta (through the xAI API; both low and high reasoning settings), as well as OpenAI's gpt-oss-120b \citep{openai2025gptoss} (through DeepInfra; low, medium, and high reasoning settings) as a recently released model with good reasoning benchmark performance that exposes its chain of thought. As in the graph coloring experiments, we run all models at temperature 0, except for DeepSeek-R1, which we run at temperature 0.5 following its developers' advice.

\subsection{Prompting and Response Evaluation}

We prompt each model (on each reasoning setting, where applicable) with each of our 2,000 problems once. The frame of assigning students to dorm rooms is the only frame in which we present the problems. We give the students alphabetically sequential male names (Alan, Bob, Charlie, etc.) and present their ratings for each other in a square matrix, specifying that each row shows one student's ratings for all of the others. (See Appendix \ref{app:stable_matching_prompts} for example prompts.) As in the graph coloring experiments, when evaluating model responses for answer correctness, we use an automatic parser and manually review and interpret any responses in which it does not detect a coherent answer.

\subsection{Results: Basic Error Patterns}

\begin{figure}[t]
\begin{centering}
\includegraphics[width=\textwidth]{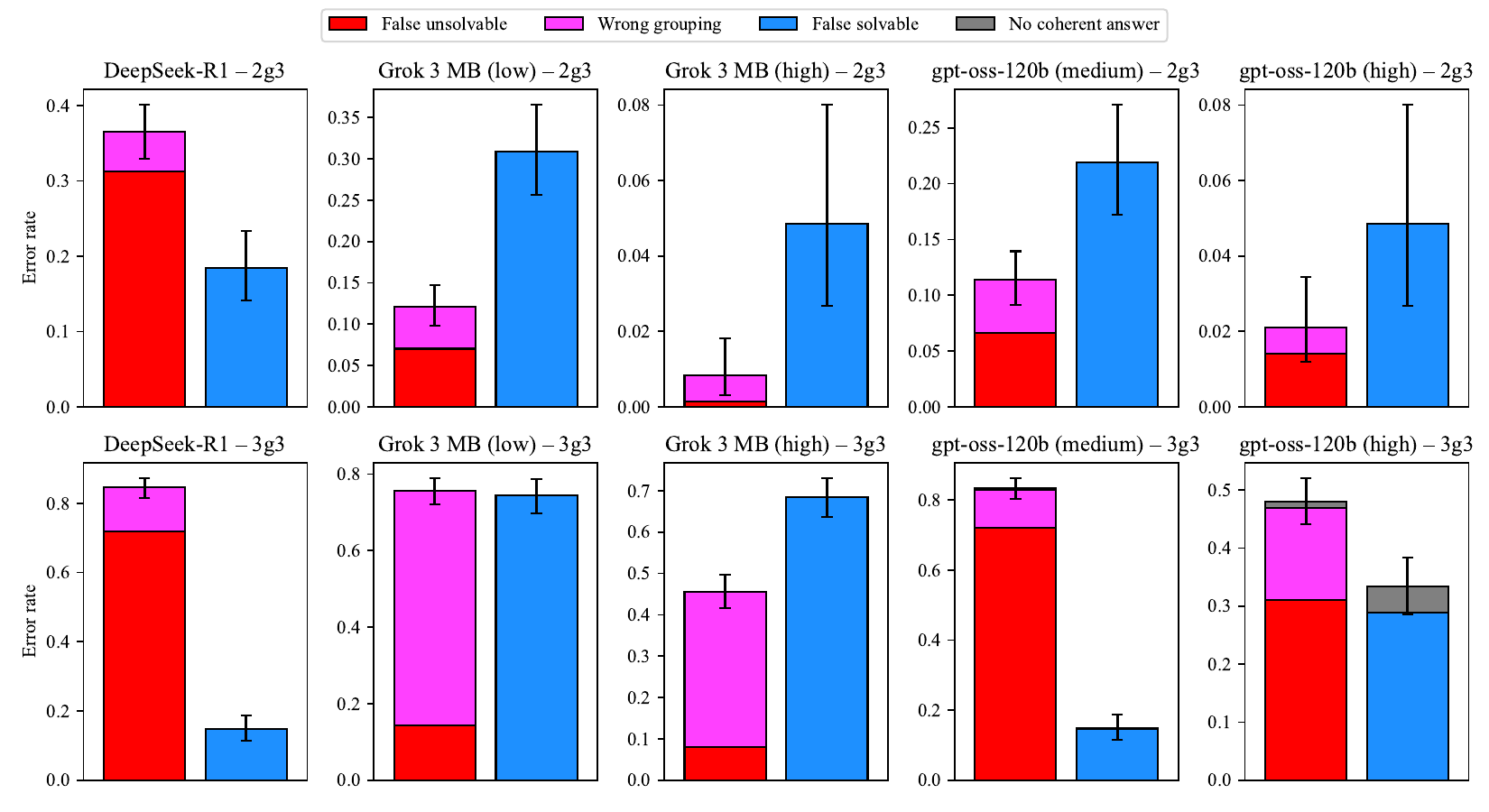}
\caption{Rate of each error type for each stable matching problem set and model. Error types are grouped by whether they apply to solvable or unsolvable problems (left and right of each pair of bars, respectively), and error rates are calculated as a fraction of the trials on the corresponding problem type. Note that each subplot's y-axis is independently scaled to its bar heights.}
\label{fig:sm_error_types}
\end{centering}
\vspace{-0.1in}
\end{figure}

Analogously to our graph coloring problems, errors on our stable 3-matching problems can be divided into three different types: claiming that a problem has no stable groupings when it actually has at least one (\textit{false unsolvable}), giving a grouping that is actually unstable for a problem with other groupings that are stable (\textit{wrong grouping}), and giving a grouping that is actually unstable for a problem with no stable groupings (\textit{false solvable}). Rates of these error types for each problem set and model are plotted in Figure \ref{fig:sm_error_types}. (The ``No coherent answer" responses from gpt-oss-120b (high) on 3g3 mostly consist of cases where it abruptly stopped talking about stable matching and ``forgot the question" partway through its chain of thought.) gpt-oss-120b (low) answered ``no stable groupings" in about $98\%$ of its 2g3 responses and 3g3 responses each, and few of its responses check any possible groupings in detail before giving their final answers; its level of reasoning effort appears too low for this task, so we exclude it from Figure \ref{fig:sm_error_types} and from the remainder of this analysis.

In contrast to the errors on our graph coloring problems, which are consistently dominated by the false-uncolorable type across RLLMs at the 8v4c complexity level and above, no comparably consistent pattern emerges here despite error rates on 2g3 already being higher than those on 8v4c for each model tested on both. The dominance of false-uncolorable errors in graph coloring appears to primarily result from edge hallucination, but with stable 3-matching there seems to be no similarly impactful reasoning flaw shared across models that biases their distribution of error types in a particular direction. As expected, error rates for each model are higher overall on 3g3 than on 2g3, and error rates on each problem set are lower overall for Grok 3 MB (high) and gpt-oss-120b (high) than for their respective lower-effort settings -- though gpt-oss-120b's decrease in false-unsolvable errors on 3g3 is counterbalanced by increases in the other types, and as its CoTs also become longer, it may be that the higher reasoning effort leads it to search for stable groupings for longer and thereby increases the likelihood of a false positive. Models' error type distributions also shift when complexity increases from 2g3 to 3g3, with errors on solvable problems becoming proportionally more common, though for Grok the increase is mostly from wrong-grouping errors while for other models it is mostly from false-unsolvable errors.

\subsection{Rating Hallucinations}

As with the graph coloring problem type, we conducted informal analysis of a sample of erroneous responses to the stable 3-matching problems and found cases in which the model's chain of thought made false claims about a particular student's rating for a particular other student (e.g. saying that Fred rates Dave a 2 when he actually rates him a 3). We want to know how common this ``rating hallucination" phenomenon is and to what extent it contributes to the errors in final answers that we observe. To this end, we use methodology similar to our graph coloring experiments, writing code to automatically parse rating statements in models' stable 3-matching responses and iteratively tuning it with priority given to eliminating false positives, thus allowing the frequencies of rating hallucinations that it finds to serve as approximate lower bounds on the true frequencies.

\subsection{Results: Rating Hallucinations}

\begin{figure}[t]
\begin{centering}
\includegraphics[width=\textwidth]{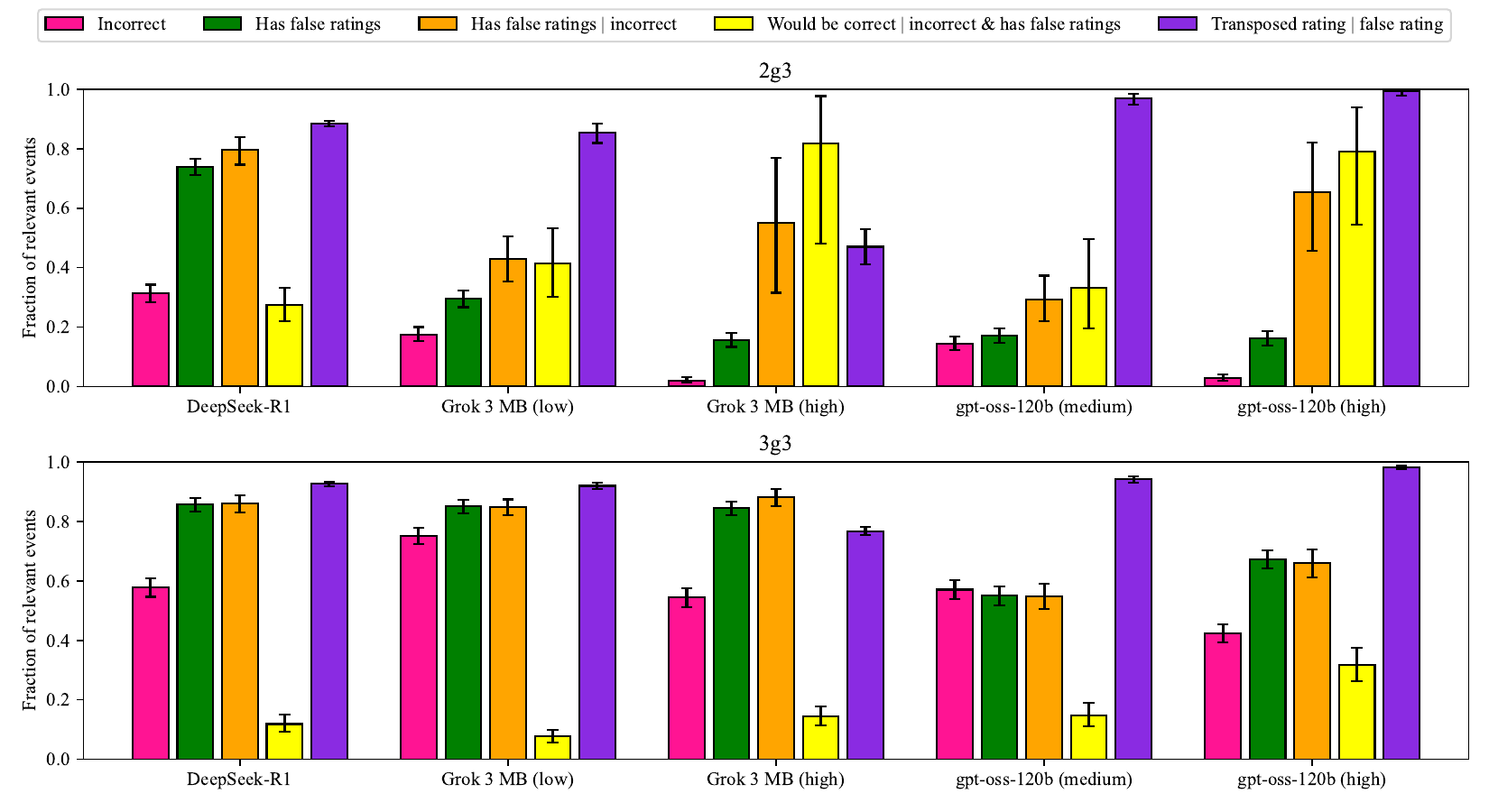}
\caption{Proportional measures related to rating hallucination for each stable matching problem set and model.}
\label{fig:sm_proportions}
\end{centering}
\vspace{-0.1in}
\end{figure}

Five proportions summarizing the rating hallucination results for each problem set and model can be seen in Figure \ref{fig:sm_proportions}. From left to right in each quintuplet of bars, these are: (1) the proportion of responses that have incorrect final answers (regardless of error type); (2) the proportion of responses that contain at least one false rating statement; (3) the proportion of responses \textit{with incorrect final answers} that contain at least one false rating statement; (4) the proportion of those responses whose final answers would be correct if the ratings matched the falsely claimed ones; and (5) the proportion of false rating statements (with repetitions of a particular statement within a particular response excluded) that would be correct if the students were transposed; i.e. \textit{A}'s rating for \textit{B} is not actually $x$, but \textit{B}'s rating for \textit{A} is $x$. We noticed the high frequency of these transposed ratings informally and chose to measure the frequency across all responses, and it appears that in almost all conditions the vast majority of false ratings are transposed ratings (with the exception of Grok 3 MB (high) on 2g3, for which the proportion of transposed ratings is still nearly $50\%$).

Leaving proportions of transposed ratings aside, the ten combinations of problem set and model appear to fall on an approximate spectrum between two extremes. One extreme is characterized by a low rate of incorrect final answers, a low rating hallucination rate, a greatly elevated rate of rating hallucinations among incorrect responses (providing at least correlational evidence by itself that rating hallucinations are contributing to error), and a high would-be-correct rate; examples include Grok 3 MB (high) on 2g3 and gpt-oss-120b (high) on 2g3. Grok 3 MB (low) on 2g3 and gpt-oss-120b (medium) on 2g3 exhibit a weaker version of this pattern: their incorrect final answer rates are higher but still below average, their rating hallucination rates are significantly but not enormously elevated among incorrect responses, and their would-be-correct rates are below $50\%$ but still above average. The other extreme is characterized by a high incorrect final answer rate, a high rating hallucination rate, an about-equally-high rating hallucination rate among incorrect responses, and a low would-be-correct rate; examples include DeepSeek-R1 on 3g3 and Grok 3 MB (low) on 3g3. In general, models on 3g3 fall toward this end of the spectrum.

These results match our graph coloring results in that overall rates of the hallucination phenomenon are higher at the higher complexity level; we also observe a pattern across the two problem types that within a given complexity level, models with lower error rates tend to have a higher proportion of their errors attributable solely to hallucinations, as if the hallucinations are more robust to conventional strategies for improving RLLM reasoning than other types of reasoning mistakes are. However, the high-error end of the spectrum of stable 3-matching conditions is unlike anything we see in our graph coloring results; the closest point of comparison might be Claude 3.7 Sonnet and Grok 3 MB (low) on 8v4c, with high rates of false-uncolorable error but low rates of error attributability to edge hallucination compared to other models. In the high-error stable 3-matching conditions (partially akin to those graph coloring ones), we think the most likely explanation for our results is that the models frequently make rating hallucinations but also frequently make other types of reasoning mistakes -- perhaps because they are less familiar with the procedure of solving this unusual problem type compared to graph coloring -- which is why most responses with rating hallucinations and incorrect final answers would still be incorrect if the ratings matched the hallucinations. However, at sufficiently low problem complexity and with a sufficiently capable model, the non-hallucination mistakes diminish even more than the hallucinations do, leaving the latter to dominate.

We find in our graph coloring experiments that the distribution of frequencies at which different edges get hallucinated is consistently non-uniform on both 8v4c and 12v6c for every model-frame combination with the relevant data available, with the edges involving the highest-index vertices tending to get hallucinated most. We do not observe equally striking patterns in the frequencies at which different ratings are hallucinated in the stable 3-matching responses; in aggregate, however, we do find weak-to-moderate but highly statistically significant biases ($p < 0.001$ via two-sided one-sample $t$-test) toward rating hallucinations involving higher-indexed students (e.g. Fred more likely than Alan) in most conditions. The only exceptions are Grok 3 MB (high) on both 2g3 and 3g3, for which the student indices involved in rating hallucinations were slightly \textit{low} on average, though neither of these biases was statistically significant ($p > 0.05$). We hypothesized that the hallucinated-edge frequency patterns in our graph coloring responses reflect weaknesses in how RLLMs represent numbers or orderings in regards to higher-indexed entities, and these results lend some support to that, though the specifics of the behavior seem to vary greatly by problem type.

Note that our findings on rating hallucination do not conflict with the previously discussed highly variable distributions of error types on these problem sets. Unlike adding an edge in a graph coloring problem, changing a rating value in a stable matching problem of this type (in a way that is not carefully targeted) does not bias the correct answer in any particular direction, since a changed rating (increased or decreased) can turn stable groupings unstable and vice versa with equal ease.

Our findings demonstrate that input-conflicting hallucinations akin to edge hallucination can contribute significantly to RLLM error on problem types other than graph coloring. The frequency of specifically \textit{transposed} ratings among false ratings across all of the models we test here may also reflect weaknesses in how RLLMs represent directionality or pair orderings (an issue that would presumably not come up in the coloring of undirected graphs), though data from more tasks would be necessary to generalize confidently.

\section{Example Prompts for Stable 3-Matching Problems} \label{app:stable_matching_prompts}

Following are two examples of the prompts presenting stable 3-matching problems that we use in the experiments we describe in Appendix \ref{app:stable_matching}, one from each of the 2g3 and 3g3 problem sets.

\subsection{2g3}

\noindent\fbox{
\parbox{\textwidth}{
\begin{allintypewriter}
Suppose there are 6 university students - Alan, Bob, Charlie, Dave, Ethan, and Fred - who must be grouped into 2 dorm rooms holding 3 students each. Suppose each student has rated how much they would want each other student as a roommate on a scale of 1 to 5, with 5 as the highest. The ratings can be found in the following matrix, with each row showing one student's ratings for all of the others:\\

\texttt{ }\texttt{ }A B C D E F\\
A X 3 5 3 5 1\\
B 5 X 1 4 4 3\\
C 4 3 X 4 4 4\\
D 1 2 5 X 5 4\\
E 2 2 2 3 X 5\\
F 1 3 4 5 5 X\\

Suppose a student's satisfaction with their assigned room is equal to the sum of their ratings for their 2 roommates. A grouping of the students into rooms is called unstable if there exists a pair of students in different rooms who could switch rooms to increase both of their satisfactions, or increase the satisfaction of one while the other's satisfaction stays the same. Do there exist any possible groupings that are stable (i.e. not unstable)? If there do not, write "Impossible" as the final line of your response. If there do, the final lines of your response should give an example of one in a format like the following:\\

Alan-Bob-Charlie\\
Dave-Ethan-Fred
\end{allintypewriter}
}
}

\newpage
\subsection{3g3}

\noindent\fbox{
\parbox{\textwidth}{
\begin{allintypewriter}
Suppose there are 9 university students - Alan, Bob, Charlie, Dave, Ethan, Fred, George, Henry, and Ian - who must be grouped into 3 dorm rooms holding 3 students each. Suppose each student has rated how much they would want each other student as a roommate on a scale of 1 to 5, with 5 as the highest. The ratings can be found in the following matrix, with each row showing one student's ratings for all of the others:\\

\texttt{ }\texttt{ }A B C D E F G H I\\
A X 2 3 4 1 5 1 1 5\\
B 5 X 2 4 5 1 4 3 2\\
C 4 2 X 4 1 5 1 1 4\\
D 2 1 5 X 3 1 5 2 5\\
E 4 3 2 3 X 2 5 1 5\\
F 1 4 2 3 1 X 4 3 5\\
G 2 1 4 4 5 5 X 2 3\\
H 5 1 5 3 3 3 2 X 3\\
I 3 4 4 5 3 2 2 5 X\\

Suppose a student's satisfaction with their assigned room is equal to the sum of their ratings for their 2 roommates. A grouping of the students into rooms is called unstable if there exists a pair of students in different rooms who could switch rooms to increase both of their satisfactions, or increase the satisfaction of one while the other's satisfaction stays the same. Do there exist any possible groupings that are stable (i.e. not unstable)? If there do not, write "Impossible" as the final line of your response. If there do, the final lines of your response should give an example of one in a format like the following:\\

Alan-Bob-Charlie\\
Dave-Ethan-Fred\\
George-Henry-Ian
\end{allintypewriter}
}
}

\end{document}